\documentclass[letterpaper, 10 pt, conference]{ieeeconf}  % Comment this line out if you need a4paper

\IEEEoverridecommandlockouts                              % This command is only needed if 
\usepackage{amsmath} % assumes amsmath package installed
\usepackage{algorithm}
\usepackage{algpseudocode}
\usepackage{graphicx}

\title{\LARGE \bf
Utilization of domain knowledge to improve POMDP belief estimation
}

\author{Tung Nguyen$^{1}$ and Johane Takeuchi$^{2}$% <-this % stops a space
\thanks{$^{1}$Tung Nguyen is with Honda Research Institute Japan
        {\tt\small tung.nguyen@jp.honda-ri.com}}%
\thanks{$^{2}$Johane Takeuchi is with Honda Research Institute Japan
        {\tt\small johane.takeuchi@jp.honda-ri.com}}%
}

\begin{document}

\maketitle
\thispagestyle{empty}
\pagestyle{empty}

%%%%%%%%%%%%%%%%%%%%%%%%%%%%%%%%%%%%%%%%%%%%%%%%%%%%%%%%%%%%%%%%%%%%%%%%%%%%%%%%
\begin{abstract}

The partially observable Markov decision process (POMDP) framework is a common approach for decision making under uncertainty. Recently, multiple studies have shown that by integrating relevant domain knowledge into POMDP belief estimation, we can improve the learned policy's performance. In this study, we propose a novel method for integrating the domain knowledge into probabilistic belief update in POMDP framework using Jeffrey's rule and normalization. We show that the domain knowledge can be utilized to reduce the data requirement and improve performance for POMDP policy learning with RL. 
\end{abstract}

%%%%%%%%%%%%%%%%%%%%%%%%%%%%%%%%%%%%%%%%%%%%%%%%%%%%%%%%%%%%%%%%%%%%%%%%%%%%%%%%
\section{INTRODUCTION}
\label{sec:introduction}
% This template provides authors with most of the formatting specifications needed for preparing electronic versions of their papers. All standard paper components have been specified for three reasons: (1) ease of use when formatting individual papers, (2) automatic compliance to electronic requirements that facilitate the concurrent or later production of electronic products, and (3) conformity of style throughout a conference proceedings. Margins, column widths, line spacing, and type styles are built-in; examples of the type styles are provided throughout this document and are identified in italic type, within parentheses, following the example. Some components, such as multi-leveled equations, graphics, and tables are not prescribed, although the various table text styles are provided. The formatter will need to create these components, incorporating the applicable criteria that follow.
Partially observable Markov decision process (POMDP) is a probabilistic sequential decision-making framework that is commonly used for robot planning \cite{kaelbling1998planning,smallwood1973optimal,astrom1965optimal,boutilier1996computing}. After formulating a decision making problem as a POMDP, we can use a reinforcement learning (RL) algorithm to learn a policy that solves this problem. POMDP has many successful applications when applied to various real-world tasks such as navigation or medical assistant robot \cite{gobelbecker2011switching,hoey2010automated}.

Many studies showed that using information about the domain can improve the performance of the robot's policy and achieve higher task completion rate \cite{zhang2012asp+,amiri2020learning,chitnis2018integrating}. A big disadvantage of existing works is that the domain knowledge used in these works is either deterministic rules, which are represented by Answer Set Programming (ASP), or needs to be manually crafted and only applicable to a very specific type of domains. This drawback puts a strict limitation to the application of the previous studies.

In this work, we propose a novel method that utilizes additional domain information when estimating the POMDP belief. Our method works with a generic representation of domain information that can be applied to a large number of decision making tasks. Our main contributions in this paper are as follows:
\begin{itemize}
    \item The proposed method employs Jeffrey's rule \cite{jeffrey1990logic} and normalization, which combines the advantages of previous studies.
    \item We demonstrate that when using the proposed method the policy learning requires fewer training episodes to converge in comparison to previous works in a simulation object fetching task. Furthermore, the learned policy that uses our proposed method achieves better performance compared to the policies learned by the previous methods.
\end{itemize}

\section{Related works}
\label{sec:related_works}
There are multiple studies attempted to integrate additional knowledge about the domain into the POMDP belief estimation to improve the belief estimation step of a POMDP. 

Zhang et al. \cite{zhang2012asp+} is one of the first studies in this line of research. In this work, the authors used the domain knowledge represented as ASP rules and use it to revise the POMDP belief state in the object localization task. The knowledge is used to determine a \emph{bias belief state} using Fechner's law. This bias belief is combined with the standard belief distribution using linear and logarithmic normalization (r-norm). \cite{zhang2012asp+} showed that the proposed method helps the robot to locate the target object more accurately. However, the method requires that the knowledge is deterministic. In addition, Fechner's law is also  applicable to a few specific domains. 

\cite{zhang2015corpp} proposed a method that uses \emph{probabilistic logic (P-log)}, which is an extension of ASP that allows logical reasoning with probabilistic rules . Similar to ASP in the method from \cite{zhang2012asp+}, P-log is used to determine a \emph{prior belief state} from the probabilistic rules. \cite{zhang2015corpp} uses the prior belief as the initialized belief state in the beginning of the planning process. The method proposed in \cite{zhang2015corpp} still needs the domain information to be manually crafted and carefully designed beforehand, which is time-consuming and not always available.

Going with a different direction, Chitnis et al. \cite{chitnis2018integrating} utilized Jeffrey's rule to revise the distributions in the belief state with``rules" created from domain knowledge. In addition, the author of this work proposed to use \emph{factored belief} to reduce the computational complexity of belief state estimation. Experiment results showed that the proposed method using Jeffrey's rule and factored belief helps improve the task completion rate. 

\section{Problem setting}
\label{sec:problem_setting}

In this section, we describe our problem setting under a formal POMDP framework and the representation of the domain knowledge. In this work, we use a factorized belief representation for the belief state, which is similar to \cite{ong2009pomdps,roy2005finding}.
\subsection{POMDP formulation} 
\label{ssec:pomdp_formulation}
Our POMDP formulation is defined as a tuple of $(S, O, B, T, Z, R)$, which are defined as follows:
\begin{itemize}
    \item $S$ - the state space. In our work, we consider domains where the a state consists of multiple attributes, $s=(f_1, f_2, ...f_N) \forall s \in S$. Each attribute $f_i$ is a discrete random variable that takes values from $\mathcal{F}_i=\{F_{i1}, F_{i2},...F_{iM_i}\}$. Here, we denote the size of each attribute's value space $\mathcal{F}_i$ as $|\mathcal{F}_i|=M_i$.
    \item $A$ - the action space, which is the set of all available actions that can be performed in the domain.
    \item $O=(O_1, O_2,...O_N)$ - the observation spaces for each of the $N$ attributes in a state. Each $O_i$ is the set of all observations for each attribute $f_i$. In our problem setting, the observation space is identical to the state space, thus, $O_i\equiv\mathcal{F}_i$ and $O\equiv\mathcal{S}$. 
    \item $B$ - the belief space. In a standard setting, the belief $b$ is a distribution over the state space $S$, which means $b=(P(s_1), P(s_2),...,P(s_i),...), s_i \in S$. However, the size of $S$ is which can be extremely large. Therefore, we employ a factorization strategy, with the belief $b$ consists of distributions of the attributes. In other words, $b = (b_1, b_2, ... b_N)$, with $b_i=(P(f_i=F_{i1}), P(f_i=F_{i2}),...P(f_i=F_{iM_{i}})), \forall i \in \{1,2..N\}$. Each $b_i$ can be seen as a distribution for random variable $f_i$, and the factorized belief uses marginalized distribution of each attribute, instead of the joint distribution.
    \item $T$ - transition function, which defines the probability of "moving" to a new state from the current state. Denote the current time step as $t$ and the current state as $s^t = (f_1^t, f_2^t,...f_N^t)$. The transition function $T$ is defined as $T(f_i^t,a^{t-1},f_i^{t-1})=(f_i^t|f_i^{t-1}, a^{t-1})$ with $a^{t-1}$ is the action chosen in the previous time step.
    \item $Z$ - observation function, which defines the probability of receiving the observation. This probability is define as $Z(o_i^t,a^{t-1},f_i^t)=P(o_i^t|f_i^t, a^{t-1})$. 
    \item $R$ - reward function, which defines the reward that is received when performing an action $a$ given state $s$, $R(s,a) = \sum_{r\in\mathcal{R}}rP(r|s,a)$. The solution of the decision making problem is a policy that maximizes the total reward that we can receive in an episode.
\end{itemize}
Let us denote the observation at time step $t-1$ as $o^{t-1}=(o_1^{t-1}, o_2^{t-1},...o_N^{t-1})$ and the current state is $s^{t-1}=(f_1^{t-1}, f_2^{t-1},...f_N^{t-1})$. The accumulation of observations and actions from the beginning of an episode to the current time step $t-1$ is called a \emph{history}, denoted by $h^{t-1}=(o^1,a^1,o^2,a^2,...o^{t-1}, a^{t-1})$. The belief consists of distributions conditioned on the history, thus, we have $b_i^{t-1}=(P(f_i^t=F_{i1}|h^{t-1}), P(f_i^t=F_{i2}|h^{t-1}),...P(f_i^t=F_{iM_{i}}|h^{t-1})), \forall i \in \{1,2..N\}$. At time step $t-1$, we perform an action $a^{t-1}$, move to next time step $t$ and observe new observation $o^t$. Next, we update our belief state $b^t$ using the following belief estimation formula,
\begin{equation}
    b_i^t=\frac{Z(o_i^t,a^{t-1},f_i^t)\sum_{j=1}^NT(f_i^t,a^{t-1},f_i^{t-1})b_j^{t-1}}{X}
    \label{eq1}
\end{equation}
, with $X$ is the normalization factor to ensure sum of the probabilities equal to 1, $b_i^t$ is the belief for attribute $i$.

\subsection{Domain knowledge}
\label{ssec:domain_knowledge}
Domain knowledge or domain information contains knowledge about the domain that we can use to revise the belief state. As mentioned above, a big disadvantage of previous works is that they all require very specific type of domain knowledge which does not generalize well to different domains. In our work, the domain knowledge is the conditional probabilities $P(f_i|f_j)$, which represents the relation between two attributes $f_i,f_j$. This type of representation is domain-agnostic and applicable to any task that have the state consists of discrete attributes. Finally, let us denote the set of domain knowledge as $\mathcal{P}$.

\subsection{Jeffrey's rule of conditioning}
\label{ssec:jeffrey_rule}
Jeffrey's rule \cite{jeffrey1990logic} of conditioning provides a way to calculate probabilities given new evidence. Let us assumes that we have a partition $\{E_1, E_2,...,E_n\}$ of an event $E$ and all the elements in this partition are mutually exclusive and exhaustive. Assume that the new probabilities $\{P^*(E_1), P^*(E_2),...,P^*(E_n)\}$ are given as evidence; with any event $A$, the new probability of $A$ can be calculated by,
\begin{equation}
    P^*(A) = \sum_{i=1}^nP(A|E_i)P^*(E_i)
    \label{eq2}
\end{equation}
Jeffrey's rule is equivalent to the judgment that the ``J-condition''
\begin{equation}
    P^*(A|E_i) = P(A|E_i)
    \label{eq3}
\end{equation}
holds for all $A$ and $E_i$. The following example demonstrates a way of applying Jeffrey's rule. Given three events $A, B,$ and $C$ with  $P(A)=0.2$, $P(B)=0.3$, and $P(A)=0.5$. Let us assume that we have a new evidence that $P^*(A) = 0.4$. Thus, we have, $P^*(B) + P^*(C) = P(\neg A)=0.6$. Therefore, $P^*(B)=0.24$ and $P^*(C)=0.36$. 
%             &= 0.6 $
% \begin{equation}
% \begin{aligned}
%   P^*(B) + P^*(C)    &= P(\neg A)\\
%             &= 0.6 \\
% \end{aligned}
% \label{eq4}
% \end{equation}
\section{Method}
\label{sec:method}
In this section, we describe our belief update method that uses domain knowledge for belief estimation. In principle, the proposed method utilizes normalization and Jeffrey's rule to revise the belief state.

\subsection{Belief normalization with bias}
\label{ssec:belief_normalization}
Our normalization method follows the same principle as in \cite{zhang2012asp+}, which calculates a bias belief from the domain information and performs normalization on the belief using the bias. However, we calculate the bias belief with chain rule given the conditional probabilities that are our domain information.

Let us consider two attributes $x$ and $y$ with support $\mathcal{X}$ and $\mathcal{Y}$, respectively. At time step $t$, we calculate the components $b_x^t, b_y^t$ of the belief state using the standard formula in Equation \ref{eq1}. Denote $P^{yx}$ as the knowledge matrix that contains information of $P(y|x)$, which satisfies, 

\begin{equation}
    P^{yx}[i][j] = P(y=j|x=i)
    \label{eq5}
\end{equation}
Similarly, we define a knowledge matrix $P^{xy}$ for $P(x|y)$. Let us recall that $b_x$ and $b_y$ can be viewed as column vectors. From the chain rule of probabilities, we have, 
\begin{equation}
\begin{aligned}
    b_x^T   &= P^{xy} \times b_y^T \\
            &= P^{xy} \times (P^{yx} \times b_x^T) \\
            &= (P^{xy} \times P^{yx}) \times b_x^T
\end{aligned}
\label{eq6}
\end{equation}

With $P^{xy}$ and $P^{yx}$ given, we can solve for $b_x$ that satisfies Equation \ref{eq6} and derive $b_y$ from it. Note that $b_x$ and $b_y$ can be seen as belief when having no observation, thus we can use them as initial belief in the beginning of each episode \cite{zhang2015corpp}. 
% In addition, even in the case we don't have one of the knowledge, for example we hae $P^{xy}$ but $P^{yx}$ is not known; we can assume $P^{yx}$ has uniform distribution without loss of generality and solves Equation \ref{eq6} normally. 
After obtaining the bias belief $b_x^*$, we integrate it into the standard belief using the following formula:

\begin{equation}
    \hat{b}_x^t = ((1-\beta)\times (b_x^t)^r + \beta \times (b_x^*)^r)^{\frac{1}{r}}  
    \label{eq7}
\end{equation}
with $r\in R$ and $\beta\in (0,1)$ are hyper parameters. Intuitively, this normalization method provides a way to ``regularize" the belief to avoid being over-confident in our estimation of the belief state, which happens in domains with noisy observation.

\subsection{Belief revision with Jeffrey's rule}
Chitnis et al. \cite{chitnis2018integrating} proposed the using of Jeffrey's rule to update the belief state with domain knowledge. Our work also utilized Jeffrey's rule but without the assumption of independence between the attributes. Let us consider $x$ and $y$ as two attributes of the state in our domain, and with the domain knowledge $P(x|y)$ and $P(y|x)$. Our belief revision algorithm is as follow:

\begin{algorithm}
    \caption{Belief revision using Jeffrey's rule}
    \label{algo2}
    \begin{algorithmic}[1] % The number tells where the line numbering should start
        \Procedure{JeffreyRevision}{$b_x^*, b_x^t$}
            \State $i \gets argmax(|b_x^*-b_x|)$
            \If {$max(|b_x^*-b_x|) > threshold$}
                \State $\hat{b}_x^t[i] = ((1-\beta)\times (b_x^t[i])^r + \beta \times (b_x^*[i])^r)^{\frac{1}{r}}$
                \State re-scale the belief $b_x^t$ with new $\hat{b}_x^t[i]$, obtain $\hat{b}_x^t$
            \Else 
                \State $\hat{b}_x^t =  b_x^t$
            \EndIf
            \State \textbf{return $\hat{b}_x^t$}
        \EndProcedure
    \end{algorithmic}
\end{algorithm}

% Note that, in Algorithm \ref{algo2}, we consider the distributions $b_x^*$, $b_y^*$, $b_x$, and $b_y$ as list of numbers to use the $max$ and $argmax$ operation. 
Note that, $threshold$ is a hyper parameter that is dependent on each domain. 
% This algorithm performs normalization (step 4) but only at the point where the difference between the normal belief $b_x$ and the bias belief $b_x^*$ is the maximum. In step 5, we apply Jeffrey's rule to obtain the new belief $\hat{b}_x^t$ by re-scaling the remaining probabilities given the new evidence, which is the probability $\hat{b}_x^t[i]$.
\subsection{Belief revision with Jeffrey's rule and normalization}
Our proposed method of integrating domain knowledge into the belief state can be viewed as a combination of Jeffrey's rule and normalization. We apply this algorithm at each turns of the interaction process.
Finally, the full interaction process is described in Algorithm \ref{algo4}.

\begin{algorithm}
    \caption{Interaction process with belief revision}
    \label{algo4}
    \begin{algorithmic}[1] % The number tells where the line numbering should start
        \Procedure{InteractionProcess}{$\mathcal{P}$} \Comment{$\mathcal{P}$ is the set of domain knowledge}
            \State calculate bias $b_x^*$ for all attribute $x$ using Equation \ref{eq6}
            \State $t \gets 0$
            \Repeat 
                \For{all attribute $x$ in state $s$}
                    \State calculate belief $b_x^t$ using Equation \ref{eq1}
                    \State  $\hat{b}_x^t \gets $\textsc{JeffreyRevision}($b^*, b_x^t$)
                    \State calculate new belief $\hat{b}_x^t$  using Equation \ref{eq7}
                \EndFor
                \State selects new action $a^t$ using policy $\pi$
                \State $t\gets t+1$
            \Until {s is terminal}
        \EndProcedure
    \end{algorithmic}
\end{algorithm}
\section{Evaluation}
\label{sec:evaluation}
This section describes the details of experiments to evaluate the proposed method of belief state estimation with domain information. 
% In particular, we assert the following two hypotheses:
% \begin{itemize}
%     \item Using domain knowledge for belief state estimation helps to improve the process of learning an RL policy that solves the given task.
%     \item The proposed method is more effective than previous methods in terms of improving RL policy learning. Additionally, the policies trained using the proposed method have better performance in comparison to the policies trained using the previous methods.
%     \end{itemize}
We conduct experiments in a simulated object fetching task, to assert these hypotheses.
% \subsection{Object fetching task}
% \label{ssec:object_fetching}
\begin{figure}[h]
      \centering
      \includegraphics[width=0.5\textwidth]{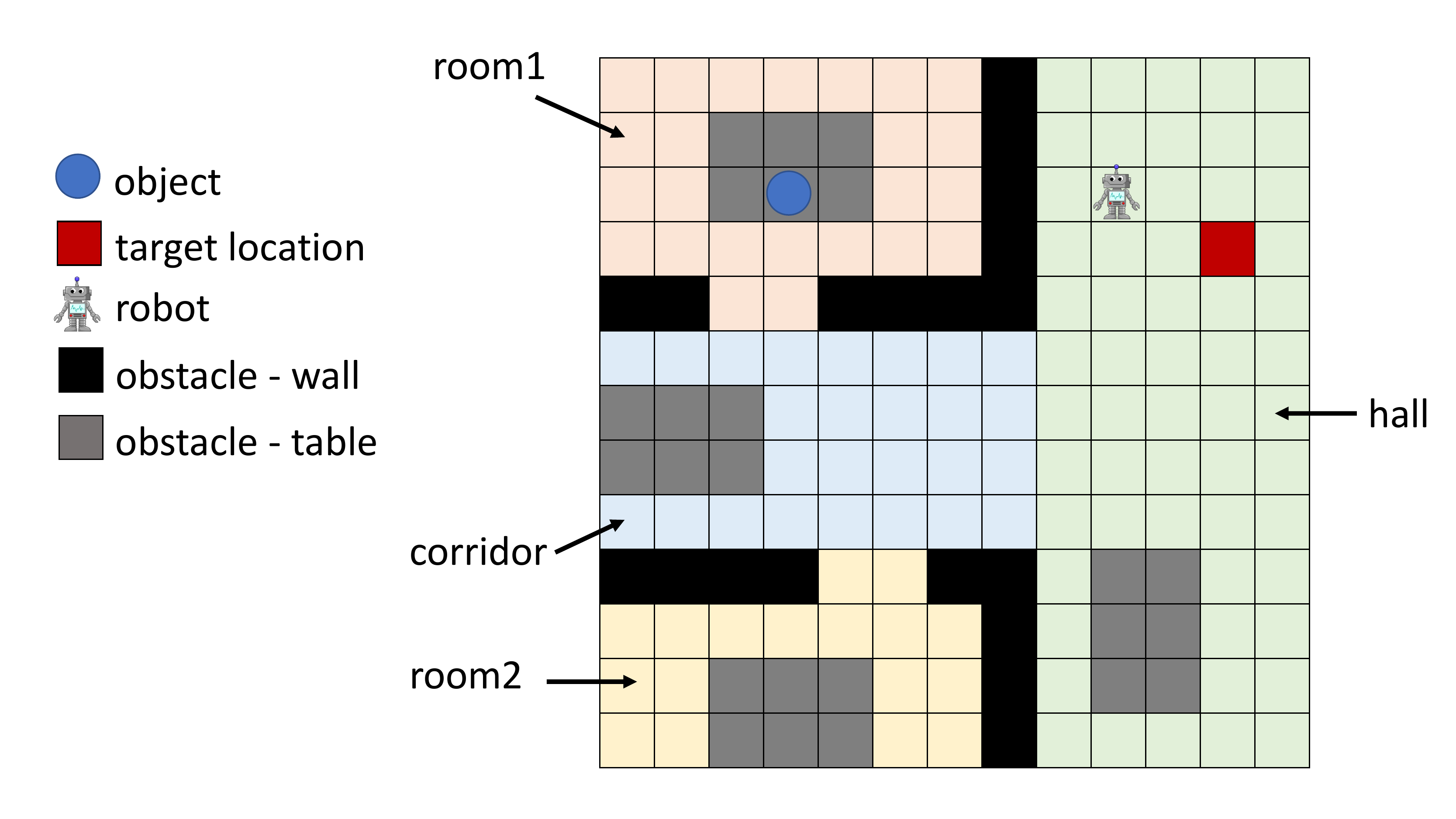}
      \caption{The object fetching task in grid-like domain with different areas.}
      \label{fig:object_fetch}
\end{figure}
\begin{figure*}[thb]
      \centering
      \includegraphics[width=0.75\textwidth]{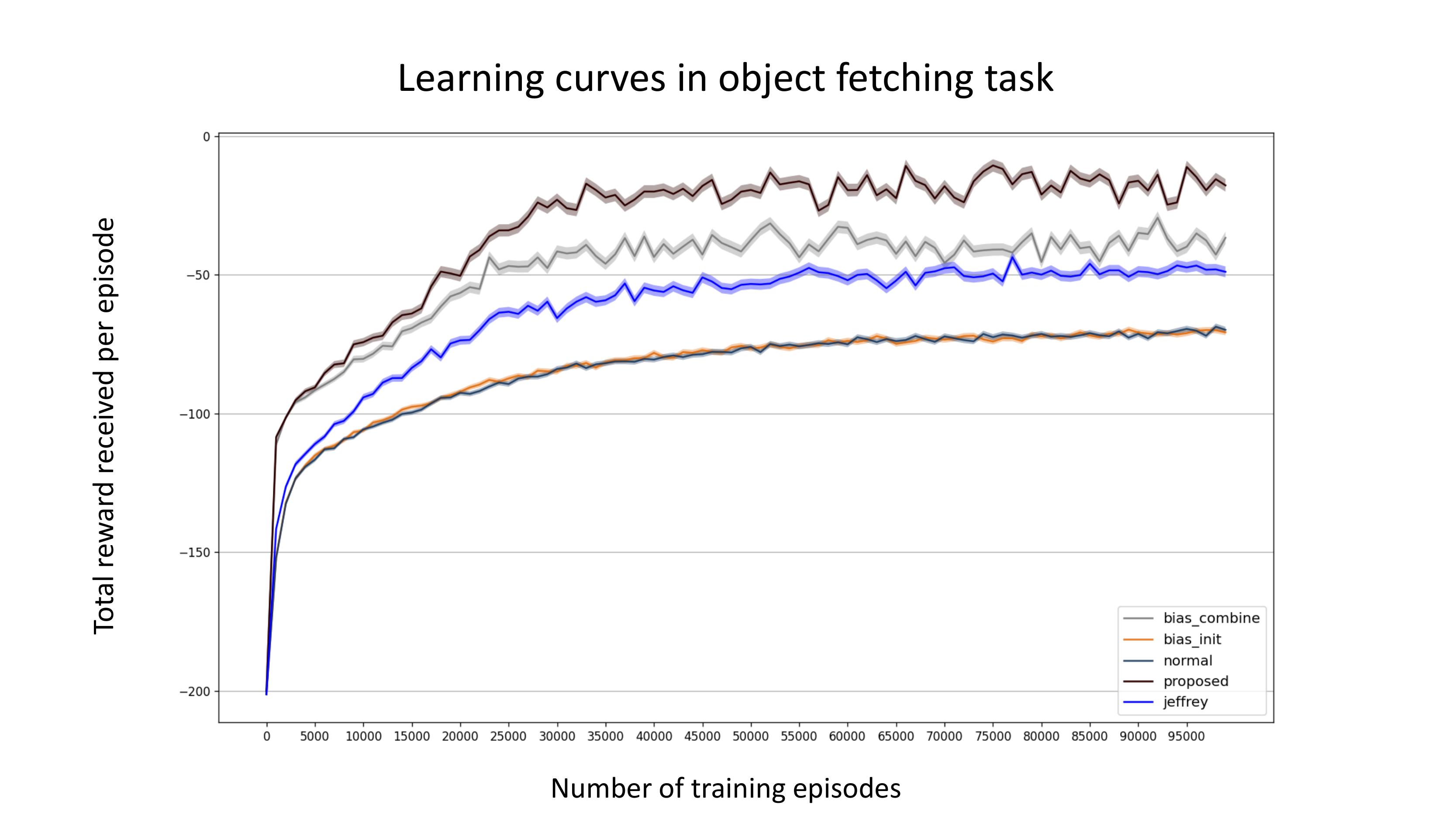}
      \caption{Experiment results in object fetching.}
      \label{fig:object_fetching_experiment}
    %   \vspace{-10pt}
\end{figure*}

\textbf{Domain and task description.} The second task is object fetching in a grid-like domain with size of $13\times 13$, as shown in Figure \ref{fig:object_fetch}. The objective of the robot in this task is to navigate through the domain to find the object (blue circle) and bring it to the target location (red). The domain is divided into four areas: room1 (pink), corridor (blue), room2 (yellow), and the hall (green). The state is represented by $s=(x,y,l,d,h)$, where $x$ and $y$ are attributes for the row and column position of the robot in the grid world, $x,y \in [0,12]$. $l$ refers to the area where the robot is at the current time step. $l$ takes value from the set $\{0 - room1, 1 - corridor, 2 - room2, 3 - hall\}$. $d$ refers to the direction which the robot is facing, $d\in\{North, East, South, West\}$. $h$ is the attribute that represents whether the robot is holding the object or not, it takes the value of $True$ or $False$. In this domain, we set $d$ and $h$ to be fully observable.

There are four actions: $TurnLeft, TurnRight, Move,$ and $Grab$. $TurnLeft$ ($TurnRight$) means the robot turns to its left (right) hand side while staying in the same position. Taking the $Move$ action moves the robot forward to the grid in front of it. When the robot performs $Grab$, it tries to grab the object in the front grid. In this domain, the robot observes $o_x, o_y,$ and $o_l$ from the three attributes $x,y,l$ that are partially observable. Let us denote the observation of $x, y, l$ as $o_x,o_y, o_l$, respectively. The observation $o_x$ is within the range $[x-2, x+2]$. The observation function is defined by,
\begin{equation}
    O(o_x, a, x)=
    \begin{cases}
        p, & \text{if $o_x=x$}\\
        (1-p)/(K-1), & \text{otherwise}
    \end{cases}
    \label{eq8}
\end{equation}  

with $p=0.3$, and $K$ is the number of elements in the set $[x-2,x+2]\cap [0,12]$. For observation of area $l$, $K$ is the number of elements in $[x-2,x+2]\cap [0,3]$. The reward function for the object fetching task is define as follow:
\begin{equation}
    R=
    \begin{cases}
        -10, & \text{if hit obstacles/boundaries}\\
        100& \text{if reach target while holding object}\\
        20& \text{if successfully find and grab the object}\\
        -1, & \text{otherwise}
    \end{cases}
    \label{eq10}
\end{equation}

We use the same type of domain knowledge as in the navigation task, which represents the relationship between two attributes $x$ and $y$. In addition, the knowledge of $P(x|l), P(l|x), P(y|l), $ and $P(l|y)$ are also used.

\textbf{Experiment setup.} In this experiment, we use Q-learning, a popular reinforcement learning algorithm, to train the policy for robot planning \cite{watkins1992q}. In order to apply Q-learning, the probabilities in the belief state are quantized to the nearest 0.1 value. The discount rate  $\lambda$ is set to 1, and learning rate $\alpha$ is 0.1. We set $\alpha$ to be multiplied by a factor of 0.9 every 500 episodes of training. $\beta$ in Equation \ref{eq4} is set to 0.5, and $r$ is set to 1. The maximum number of steps the robot can take during one episode is 200. Due to differences in problem and formulation, it is not possible to directly use the methods from previous studies. However, we implemented baselines based on these studies for comparison with our proposed method. The following models are used in this experiment:
\begin{itemize}
    \item \emph{normal:} the method that does not use domain knowledge and only use the standard formula for belief estimation.
    \item \emph{bias\_combine:} this method is from \cite{zhang2012asp+}, which uses Equation \ref{eq7} for belief revision.
    \item \emph{bias\_init:} this method is inspired by the work in \cite{amiri2020learning}, which uses the bias belief for initialization at the beginning of each episode.
    \item \emph{jeffrey:} this method is based on the idea from \cite{chitnis2018integrating}, which uses Jeffrey's rule.
    \item \emph{proposed:} the proposed method
\end{itemize}
% We had to leave the \emph{jeffrey} baseline out due to incompatibilities with this task. 

\textbf{Experiment results.} From Figure \ref{fig:object_fetching_experiment}, we can see that using domain knowledge for belief revision significantly improves the convergence of the training process, and the model converges the fastest when using our proposed method.

Table \ref{tab:results_object_fetch} shows the performance of the policies trained by different methods. Similar to the results above, we can see that the robot learns to solve the task with higher success rate when using the domain knowledge for belief revision. In addition, the policy trained by the proposed method achieves the best performance and significantly outperforms the previous methods in both metrics. 

\begin{table}[h]
\caption{Performance of trained policies in object fetching}
\label{tab:results_object_fetch}
\begin{center}
\begin{tabular}{|c||c|c|}
\hline
Models & Average total reward per episode & Success rate\\
\hline
Normal & -36.98 $\pm$ 2.24 & 6.58\%\\
\hline
Bias Init & -70.76 $\pm$ 1.07 & 5.86\%\\
\hline
Bias Combine & -36.98 $\pm$ 2.24 & 40.00\%\\
\hline
Jeffrey & -48.19 $\pm$ 1.96 & 26.10\%\\
\hline
Proposed & -17.67 $\pm$ 2.27 & 52.48\%\\
\hline
\end{tabular}
\end{center}
\end{table}
\section{CONCLUSIONS}

In this study, we propose a novel method that uses additional domain knowledge for POMDP belief estimation process. Our proposed method use normalization and Jeffrey's rule of conditioning to revise the belief state. We demonstrated that the proposed method helps to improve policy learning of the robot in two simulation domains and achieve significantly better result in comparison to previous methods. 

In the future, we would like to conduct experiment with physical environment to further confirm the effectiveness of the proposed method. 

\addtolength{\textheight}{-12cm}   % This command serves to balance the column lengths

\bibliography{mybib}

\begin{thebibliography}{10}
\providecommand{\url}[1]{#1}
\csname url@rmstyle\endcsname
\providecommand{\newblock}{\relax}
\providecommand{\bibinfo}[2]{#2}
\providecommand\BIBentrySTDinterwordspacing{\spaceskip=0pt\relax}
\providecommand\BIBentryALTinterwordstretchfactor{4}
\providecommand\BIBentryALTinterwordspacing{\spaceskip=\fontdimen2\font plus
\BIBentryALTinterwordstretchfactor\fontdimen3\font minus
  \fontdimen4\font\relax}
\providecommand\BIBforeignlanguage[2]{{%
\expandafter\ifx\csname l@#1\endcsname\relax
\typeout{** WARNING: IEEEtran.bst: No hyphenation pattern has been}%
\typeout{** loaded for the language `#1'. Using the pattern for}%
\typeout{** the default language instead.}%
\else
\language=\csname l@#1\endcsname
\fi
#2}}

\bibitem{kaelbling1998planning}
L.~P. Kaelbling, M.~L. Littman, and A.~R. Cassandra, ``Planning and acting in
  partially observable stochastic domains,'' \emph{Artificial intelligence},
  vol. 101, no. 1-2, pp. 99--134, 1998.

\bibitem{smallwood1973optimal}
R.~D. Smallwood and E.~J. Sondik, ``The optimal control of partially observable
  markov processes over a finite horizon,'' \emph{Operations research},
  vol.~21, no.~5, pp. 1071--1088, 1973.

\bibitem{astrom1965optimal}
K.~J. Astrom, ``Optimal control of markov decision processes with incomplete
  state estimation,'' \emph{J. Math. Anal. Applic.}, vol.~10, pp. 174--205,
  1965.

\bibitem{boutilier1996computing}
C.~Boutilier and D.~Poole, ``Computing optimal policies for partially
  observable decision processes using compact representations,'' in
  \emph{Proceedings of the National Conference on Artificial
  Intelligence}.\hskip 1em plus 0.5em minus 0.4em\relax Citeseer, 1996, pp.
  1168--1175.

\bibitem{gobelbecker2011switching}
M.~G{\"o}belbecker, C.~Gretton, and R.~Dearden, ``A switching planner for
  combined task and observation planning,'' in \emph{Twenty-Fifth AAAI
  Conference on Artificial Intelligence}, 2011.

\bibitem{hoey2010automated}
J.~Hoey, P.~Poupart, A.~von Bertoldi, T.~Craig, C.~Boutilier, and
  A.~Mihailidis, ``Automated handwashing assistance for persons with dementia
  using video and a partially observable markov decision process,''
  \emph{Computer Vision and Image Understanding}, vol. 114, no.~5, pp.
  503--519, 2010.

\bibitem{zhang2012asp+}
S.~Zhang, M.~Sridharan, and F.~S. Bao, ``Asp+ pomdp: Integrating non-monotonic
  logic programming and probabilistic planning on robots,'' in \emph{2012 IEEE
  International Conference on Development and Learning and Epigenetic Robotics
  (ICDL)}.\hskip 1em plus 0.5em minus 0.4em\relax IEEE, 2012, pp. 1--7.

\bibitem{amiri2020learning}
S.~Amiri, M.~S. Shirazi, and S.~Zhang, ``Learning and reasoning for robot
  sequential decision making under uncertainty,'' in \emph{Proceedings of the
  AAAI Conference on Artificial Intelligence}, vol.~34, no.~03, 2020, pp.
  2726--2733.

\bibitem{chitnis2018integrating}
R.~Chitnis, L.~P. Kaelbling, and T.~Lozano-P{\'e}rez, ``Integrating
  human-provided information into belief state representation using dynamic
  factorization,'' in \emph{2018 IEEE/RSJ International Conference on
  Intelligent Robots and Systems (IROS)}.\hskip 1em plus 0.5em minus
  0.4em\relax IEEE, 2018, pp. 3551--3558.

\bibitem{jeffrey1990logic}
R.~C. Jeffrey, \emph{The logic of decision}.\hskip 1em plus 0.5em minus
  0.4em\relax University of Chicago press, 1990.

\bibitem{zhang2015corpp}
S.~Zhang and P.~Stone, ``Corpp: Commonsense reasoning and probabilistic
  planning, as applied to dialog with a mobile robot,'' in \emph{Proceedings of
  the AAAI Conference on Artificial Intelligence}, vol.~29, no.~1, 2015.

\bibitem{ong2009pomdps}
S.~C. Ong, S.~W. Png, D.~Hsu, and W.~S. Lee, ``Pomdps for robotic tasks with
  mixed observability.'' in \emph{Robotics: Science and Systems}, vol.~5, 2009,
  p.~4.

\bibitem{roy2005finding}
N.~Roy, G.~Gordon, and S.~Thrun, ``Finding approximate pomdp solutions through
  belief compression,'' \emph{Journal of artificial intelligence research},
  vol.~23, pp. 1--40, 2005.

\bibitem{watkins1992q}
C.~J. Watkins and P.~Dayan, ``Q-learning,'' \emph{Machine learning}, vol.~8,
  no. 3-4, pp. 279--292, 1992.

\end{thebibliography}
\bibliographystyle{IEEEtran}

\end{document}